\documentclass[letterpaper, 10 pt, conference]{ieeeconf}  

\IEEEoverridecommandlockouts                              

\usepackage{graphics} 
\usepackage{graphicx} 
\usepackage{epsfig} 
\usepackage{mathptmx} 
\usepackage{times} 
\usepackage{amsmath} 
\usepackage{amssymb}  
\usepackage{booktabs} 
\usepackage{array}    

\usepackage[]{xcolor}
\renewcommand{\textcolor}[2]{#2}

\usepackage{tabularx}
\definecolor{Payam}{RGB}{100, 100, 255} 

\usepackage{subfigure}
\usepackage{amsmath} 
\usepackage{array}   
\usepackage[font={small}]{caption}

\usepackage{algorithm}
\usepackage{algpseudocode}
\usepackage{array}
\usepackage{comment}
\usepackage[pagebackref,breaklinks,colorlinks,citecolor=cvprblue]{hyperref}
\usepackage{cite}

\usepackage{mathrsfs}
\usepackage{amssymb}
\usepackage{gensymb}

\usepackage{subcaption}
\usepackage{graphicx}
\usepackage{lipsum}  
\usepackage{xcolor}

\title{\LARGE \bf
MotionScript: Natural Language Descriptions \\ for Expressive 3D Human Motions
}

\author{Payam Jome Yazdian$^{1}$, Rachel Lagasse$^{1}$, Hamid Mohammadi$^{2}$, Eric Liu$^{1}$, Li Cheng$^{2}$ and Angelica Lim$^{1}$
\thanks{*This work was supported by HW-SFU Joint Lab}
\thanks{$^{1}$School of Computing Science,
        Simon Fraser University, Burnaby, BC, Canada
        {\tt\small payam\_jome-yazdian@sfu.ca}}%
\thanks{$^{2}$Dept. of Electrical and Computer Engineering, University of Alberta, Edmonton, Alberta, Canada}%
}

\begin{document}
\maketitle
\thispagestyle{empty}
\pagestyle{empty}



\begin{abstract}
We introduce MotionScript, a novel framework for generating highly detailed, natural language descriptions of 3D human motions. Unlike existing motion datasets that rely on broad action labels or generic captions, MotionScript provides fine-grained, structured descriptions that capture the full complexity of human movement—including expressive actions (e.g., emotions, stylistic walking) and interactions beyond standard motion capture datasets. MotionScript serves as both a descriptive tool and a training resource for text-to-motion models, enabling the synthesis of highly realistic and diverse human motions from text. By augmenting motion datasets with MotionScript captions, we demonstrate significant improvements in out-of-distribution motion generation, allowing large language models (LLMs) to generate motions that extend beyond existing data. Additionally, MotionScript opens new applications in animation, virtual human simulation, and robotics, providing an interpretable bridge between intuitive descriptions and motion synthesis. To the best of our knowledge, this is the first attempt to systematically translate 3D motion into structured natural language without requiring training data. Code, dataset, and examples are available at \href{https://pjyazdian.github.io/MotionScript}{https://pjyazdian.github.io/MotionScript}
\end{abstract}    
\section{Introduction}
{
    \label{sec:intro}
    

    {
        In computer animation, a production artist can animate a specific motion, such as waving a hand, in many ways, from a quick flick of the right wrist to an enthusiastic left arm swing. 
        Text-to-motion algorithms that can mimic the precision, diversity, and flexibility of the animation process ~\cite{T2MGPT} are especially useful for generating motions of virtual humans for robotics simulators~\cite{biro2021sfu}, co-speech gesture generation~\cite{GENEA2023}, sign-language generation~\cite{text-to-sign}, retrieval~\cite{ReMoDiffuse}, crowd animation, and more. These methods are typically trained on paired motions and natural language annotations from the small-scale KIT Motion-Language \cite{KITML} and HumanAct12 \cite{Action2motion} datasets as well as large-scale AMASS \cite{AMASS} dataset with extended human annotations from BABEL \cite{BABEL} and HumanML3D \cite{HUMANML3D}. For text-driven motion generation tasks, a broad range of methods such as Transformers~\cite{ude, ACTOR}, GANs~\cite{ActFormer}, VAEs~\cite{TEMOS, Teach, SINC}, VQ-VAEs~\cite{T2MGPT} and diffusions \cite{tevet2023human, PhysDiff, MoFusion, Flame} have been employed. These methods can perform high quality and diverse text-to-motion synthesis.
    }

\begin{figure}[t]
        \centering
        \includegraphics[width=\columnwidth]{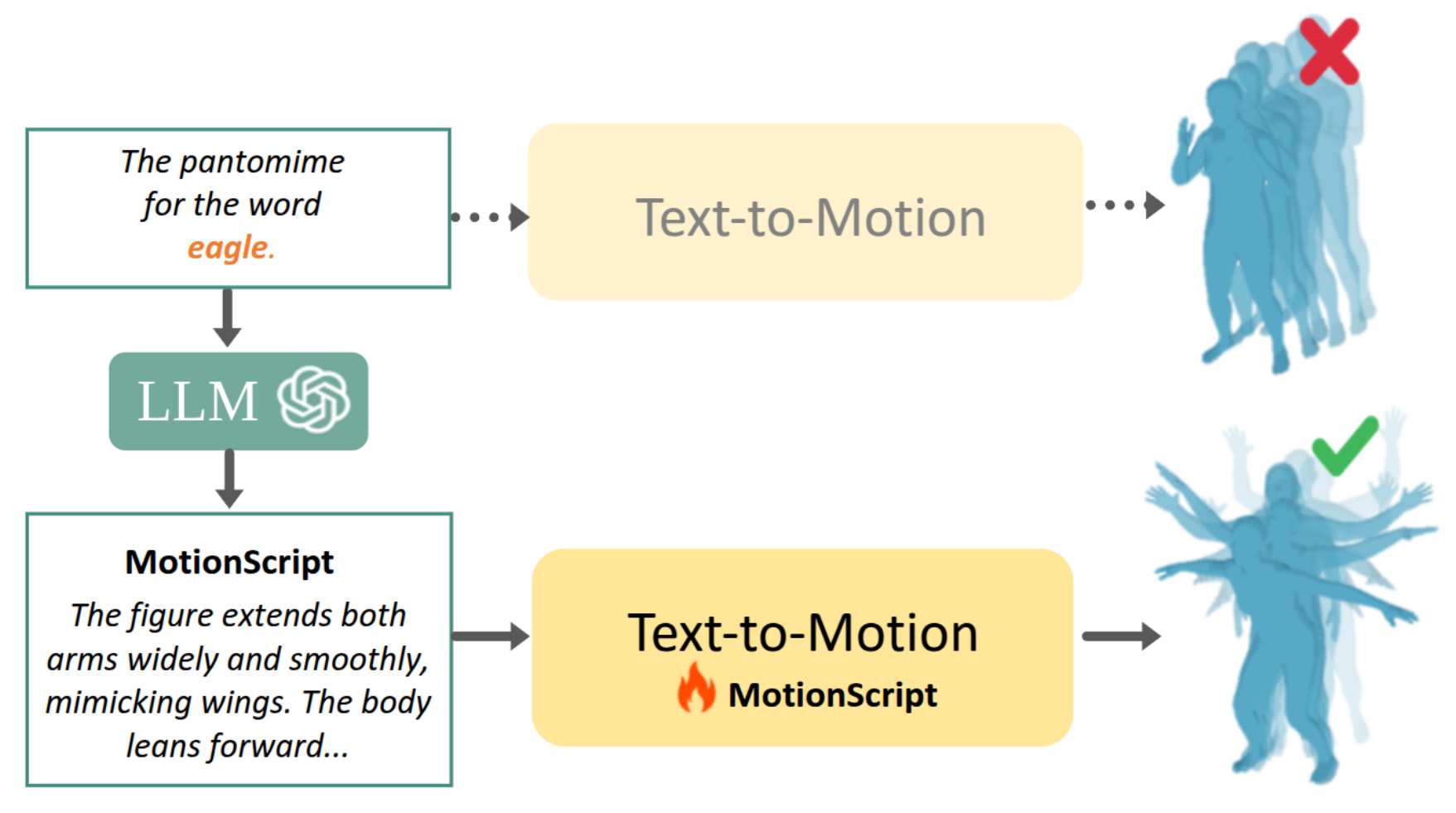}
        \caption{
        Motionscript provides a structured language for open-vocabulary, fine-grained motion descriptions. By integrating LLMs and a motionscript fine-tuned text-to-motion model, this pipeline enables out-of-distribution motion generation where standard text-to-motion models perform sub-optimally.
        }
        \vspace{-5mm}
        \label{fig:experiment_app_llm}
\end{figure}
        
    {
        Yet, current methods struggle to generate motions that were unseen in the training dataset. These out-of-distribution motions may be especially challenging because they require more high-level understanding of context and emotion, including interactions with humans, e.g. \textit{getting called on by the teacher and not knowing the answer}, animals, e.g.  \textit{reacting to a swarm of bees}, and environment, e.g. \textit{eating an ice cream that is melting really quickly}. In addition, they could include stylistic characterizations or pantomime, e.g. \textit{an elderly woman doing water aerobics} or \textit{a bullfighter in a match with a bull}. In robotics simulators, varied types of realistic actions and reactions could be needed for human-like simulations, e.g. \textit{waving down a taxi, but it passes by without stopping}. The combinatorial complexity of actions modified by context makes it challenging to rely solely on actions recorded in existing datasets.}

    }

    {
        In this paper, we propose MotionScript, a fine-grained natural language description of human body motions. MotionScript is built upon PoseScript \cite{posescript}, a method for algorithmically describing static poses in natural language, extending it to the temporal domain. \textcolor{red}{By leveraging granular MotionScript descriptions, we bridge the gap between 3D motion and LLM reasoning, unlocking their ability to generate MotionScript-like descriptions for challenging, out-of-distribution texts}, e.g. \textit{pantomime for the word eagle}, (Fig.~\ref{fig:experiment_app_llm}), and text-to-motion models trained on MotionScript captions can ultimately produce 3D human motions that are preferred by users on these out-of-distribution samples. 
    }
    
    In summary, our contributions are: 
    \begin{enumerate}
        \item \textcolor{blue}{Introducing MotionScript as a fine-grained, structured motion description framework that can be generated from motion data, human input, or LLMs. We also propose an algorithm that automatically extracts MotionScript descriptions from 3D human motion, enabling the augmentation of motion datasets with detailed, interpretable captions.}

        \item \textcolor{blue}{Training a text-to-motion model on MotionScript-augmented datasets, demonstrating that models trained with MotionScript can generate more expressive and out-of-distribution motions compared to standard text-to-motion baselines.}

        \item \textcolor{blue}{To the best of our knowledge, this is the first attempt to systematically describe 3D human motion in structured natural language, providing an interpretable and scalable motion representation without relying on predefined action categories.}
    \end{enumerate}

\section{Related Work}
\label{sec:related_work}
{

    \noindent\textbf{Human Motion Generation.}
    {
        3D human representation is essential for human-related tasks. For instance, EVA3D \cite{Eva3d} and Text2Perform \cite{Text2Performer} proposed intermediary pose representations to improve motion synthesis. Moreover, generative models for realistic human motion synthesis include motion prediction~\cite{yuan2020dlow, zhong2022spatio, liu2022towards}, motion generation \cite{tevet2023human, Ganimator}, co-speech gesture~\cite{Gesture2Vec, GestureDiffuCLIP}, dance~\cite{LISTEN_DANCE_ACTION} and sign-language generation\cite{text-to-sign}. Various modalities have been explored for conditional motion generation, including action classes~\cite{guo2020action2motion, petrovich2021action}, motion descriptions~\cite{Language2pose, Teach, Flame}, music~\cite{BRACE, AIST++}, speech~\cite{DiffuseStyleGesture, Taming}, scene context~\cite{starke2019neural, xu2023interdiff}, and style~\cite{ZeroEGGS, BEAT}.
        We mainly focus on text-conditioned motion synthesis \cite{Language2pose, Teach, ghosh2021synthesis, guo2022generating, TEMOS, PhysDiff, Flame, MoFusion, qian2023breaking, ReMoDiffuse} with a particular emphasis on data augmentation.
    }

    \noindent\textbf{Text-Driven Human Motion Generation}
    {
        Integrating linguistic descriptions with visual data allows for more realistic and human-like behavior generation. Early systems \cite{Action2motion, Posegpt} used categorical labels, while KIT ML \cite{KITML} introduced free-form language descriptions beyond fixed classes. Later on, HumanML3D~\cite{HUMANML3D} and BABEL~\cite{BABEL}  expanded the human annotations of AMASS~\cite{AMASS}, a large-scale motion dataset.
        Several deterministic methods \cite{Language2pose, ghosh2021synthesis} 
        and
        probabilistic approaches such as transformers \cite{ude, ACTOR}, GANs \cite{ActFormer}, VAEs \cite{TEMOS, Teach, SINC}, VQ-VAEs \cite{T2MGPT}, and diffusions \cite{tevet2023human, PhysDiff, MoFusion, Flame} were also proposed to advance the motion generation task.

        Alongside these generative models, it is recognized that performance in motion generation can be boosted through data augmentation. UnifiedGesture \cite{UnifiedGesture} introduced unifying diverse skeletal data to extend co-speech gesture datasets, and MCM \cite{MCM} proposed a training pipeline to integrate motion datasets, facilitating the creation of multi-condition, multi-scenario motion data such as human motion \cite{HUMANML3D}, co-speech gesture \cite{BEAT}, and dance \cite{AIST++}. AMD \cite{AMD} proposed a method to generate motions conditioned on previous time step and current text description in an autoregressive fashion to manage the scarcity of human motion-captured data for long prompts. EDGE \cite{Edge} generates arbitrarily long dances, by enforcing temporal continuity between batches of multiple sequences. Make-An-Animation (MAA) \cite{Make-An-Animation} uses a two-stage training approach: first, it extracts pose-text pairs from large-scale image-text datasets to generate diverse motions, then it fine-tunes on motion capture data to model temporal dynamics. Fg-T2M \cite{Fg-T2M} generates fine-grained human body motions by analyzing linguistic structures in motion captions, while SINC \cite{SINC} and \cite{xiang2023generative, Action-gpt} utilize LLMs \cite{GPT3} to enhance motion data by integrating detailed text descriptions and generating complex, simultaneous actions.
    }
    
    \subsection{Human Motion Semantic Representation}
    {
        Several methods generate semantic labels or captions from skeleton data. Posebits \cite{pons2014posebits} introduced binary captions describing articulation angle or relative position of joints, while \cite{pavlakos2018ordinal} used ordinal depth relations of joints as a supervisory signal sourced from human annotators. Poselets \cite{Poselets} extracts intermediate but not easy to interpret pose information considering anthropometric constraints. FixMyPose \cite{Fixmypose} and AIFit \cite{Aifit} introduced captions offering finer granularity to distinguish poses. GDL \cite{GDL} proposed a rule-based Gesture Description Language (GDL) to represent human body skeleton data with synthetic descriptions. Hierarchical Motion Understanding \cite{HMProgram} proposed a program-like representation that described motions with high-level parametric primitives i.e. circular, linear, or stationary extended by adding general spline primitives \cite{HMProgrammatic}. PoseScript \cite{posescript} recently introduced a rule-based algorithm that converts pose data into natural language descriptions, starting with lower-level representations such as \textit{`the left elbow is \textgreater $90\degree$ (bent) or \textless $90\degree$ (straight)'} and translating these into sentences using templates.
        

        Existing captioning methods either rely on human annotations or automatic captions that lack the granularity needed for fine-grained motion generation. Furthermore, augmenting captions with LLMs without a direct connection to the corresponding motion can lead to misalignments due to the many-to-many nature of text-to-motion mapping.

    }

\section{MotionScript Representation}
\label{sec:motionscript}


\begin{figure*}
    \centering
    \includegraphics[width=1.0\linewidth]{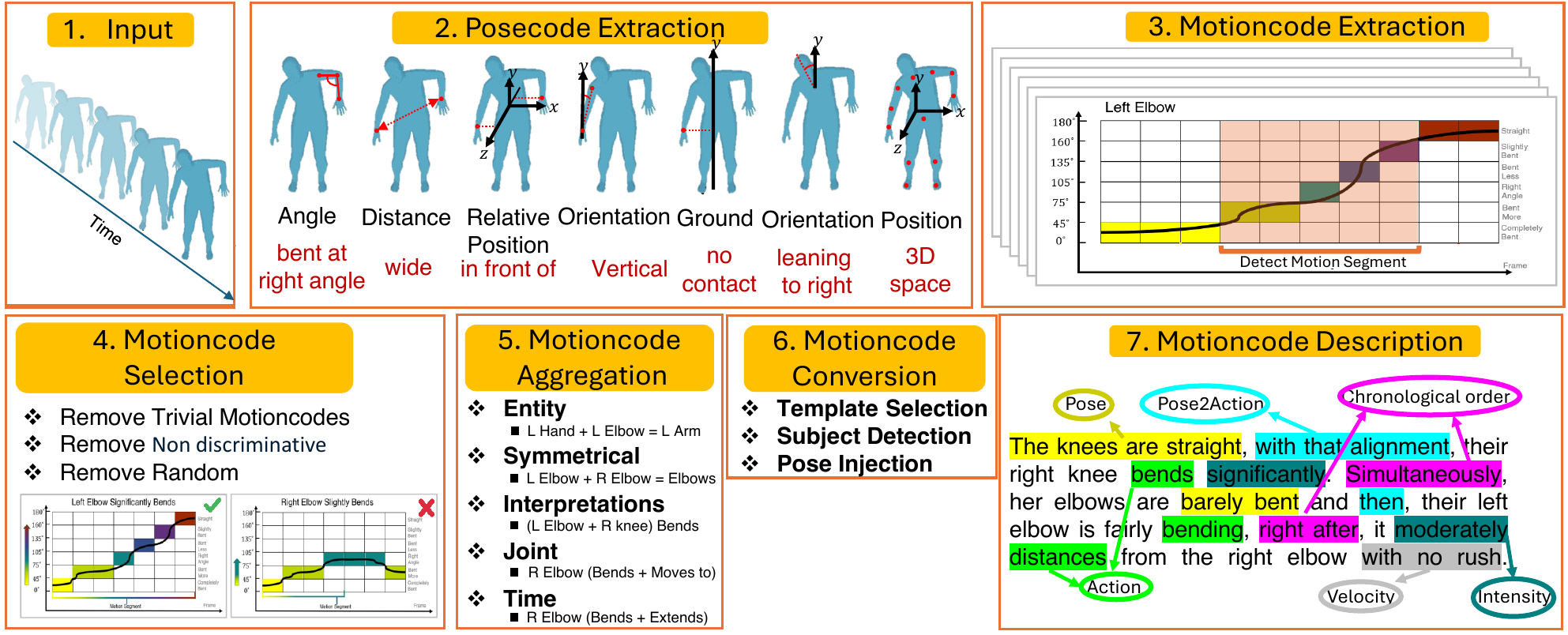}
    \caption{The proposed MotionScript framework converts a sequence of 3D poses into a sequence of posecodes, detects and selects important motions, and finally aggregates them and converts them into text.}
    \label{fig:ms_framwork}
    \vspace{-5mm}
\end{figure*}


\textcolor{blue}{In this paper, we present MotionScript, a fine-grained natural language framework for describing 3D human motion by automatically generating detailed and structured textual captions from motion data. Unlike conventional methods that rely on broad action labels or isolated pose descriptions, MotionScript captures both spatial and temporal dynamics in a comprehensive manner. We propose an algorithm that extracts MotionScript descriptions directly from 3D human joint trajectories, thereby augmenting motion datasets with interpretable and scalable captions. Furthermore, by training a text-to-motion synthesis model on MotionScript-augmented data, we demonstrate enhanced expressive power and improved generation of out-of-distribution motions compared to standard baselines. Our approach builds upon the pose-level representation introduced in \cite{posescript} and extends it by incorporating temporal attributes, as validated on a dataset combining MotionScript captions with those from the HumanML3D dataset~\cite{HUMANML3D}.}


\subsection{Automatic Motion Caption Generation}

This section explains generating textual representations from 3D skeleton sequences. As illustrated in Fig.~\ref{fig:ms_framwork}, we first extract \textit{posecodes}, a quantifiable representation of static pose attributes. We then analyze temporal changes in \textit{posecodes} categories to capture the motion dynamics. Algorithm~\ref{algorithm_motion_detection} outlines how to identify \textit{motioncodes}, a new representation of movement patterns. We then exclude redundant \textit{motioncodes} and aggregate them to produce concise and coherent natural language sentences.

As shown in Fig.~\ref{fig:caption-example}, the input is 3D joint coordinates from the SMPL-H model \cite{MANO:SIGGRAPHASIA:2017}, with default shape coefficients and a normalized global y-axis orientation relative to the first frame. The output is an English sentence.

\subsection{Posecode Extraction}

A \textit{posecode} categorizes the spatial or angular relationships between joints in a frame using predefined thresholds, such as joint distances or angles.
Prior work used six types of elementary pose information including angles, distances, relative positions \cite{pons2014posebits}, pitch, roll, and ground-contacts \cite{posescript}. 
MotionScript extends \textit{posecode} to include body orientation and 3D position, required for describing dynamic motions.

     \noindent\textbf{Angle posecodes} discretize a body part's angle, (e.g., left elbow) into: \{`straight', `slightly bent', `partially bent', `bent at a right angle', `almost completely bent', `completely bent'\}.

    \noindent\textbf{Distance posecodes} classify the L2-distance between body parts (e.g., hands) into: {`close', `shoulder width', `spread', and `wide apart'}.

    \noindent\textbf{Relative Position posecodes} explain a joint's position relative to another over the X-axis \{`right of', `left of'\}, Y-axis \{`below', `above'\}, and Z-axis \{`behind', `in front of'\}. Vague positions, i.e. near the border, are denoted as `ignored'.

    \noindent\textbf{Pitch \& Roll posecodes} describe a body part's orientation, such as the left knee and hip defining the left thigh, as `vertical' or `horizontal' relative to the y-plane, with orientations that fall between these extremes categorized as `ignored'.

    \noindent\textbf{Ground-Contact posecodes} are defined exclusively as an intermediate calculation and indicate whether a body keypoint is `on the ground' or `ground-ignored'.


    \noindent\textbf{Orientation posecodes} define the spatial orientation of the body root relative to the first frame, using three axes to determine its orientation.
    
    \noindent\textbf{Position posecodes} 
    explain both the relative position of body joints to the body root and the global position in 3D space.

The 3D joints are normalized so that the avatar starts facing forward at the origin. To account for subjectivity, we add noise to the measured angles and distances before classifying them into posecodes.

\subsection{Motioncode Extraction}

\textcolor{blue}{
A \textit{motioncode} $\mathbf{M}$ represents the dynamics of a joint movement associated with a specific \textit{posecode} $\mathbf{P}$ over time. It consists of three key attributes:
1. \textbf{Temporal Attribute ($\mathbf{M}_T$):} Defines the motion interval, starting at $\mathbf{M}_{T_s}$ and ending at $\mathbf{M}_{T_e}$.
2. \textbf{Spatial Attribute ($\mathbf{M}_S$):} Quantifies the cumulative number of transitions in the associated \textit{posecode} categories $\mathbb{C}_{\mathbf{P}}$. It is given by:
\vspace{-3mm}
   \[
   \mathbf{M}_{S} = \sum_{t=\mathbf{M}_{T_s}}^{\mathbf{M}_{T_e}-1} \Delta(\mathbb{C}_{\mathbf{P}}, t)
   \]
   where $\Delta(\mathbb{C}_{\mathbf{P}}, t)$ represents changes in \textit{posecode} categories from time $t$ to $t+1$. The magnitude $\left| {\mathbf{M}_{S}} \right|$ and the sign $\mathbf{sgn}({\mathbf{M}_{S}})$ indicate the motion's intensity and direction, respectively.
3. \textbf{Velocity Attribute ($\mathbf{M}_V$):} Measures the rate of change in the spatial attribute over time, computed as:
   \[
   \mathbf{M}_{V} = \frac{\left| {\mathbf{M}_{S}} \right| }{ \mathbf{M}_{T_e} - \mathbf{M}_{T_s}}
   \]
   Based on noise-injected thresholds, velocity is classified into five categories: `very slow', `slow', `moderate', `fast', and `very fast'.}
\begin{figure}[t]
    \centering
    \includegraphics[width=1.0\linewidth]{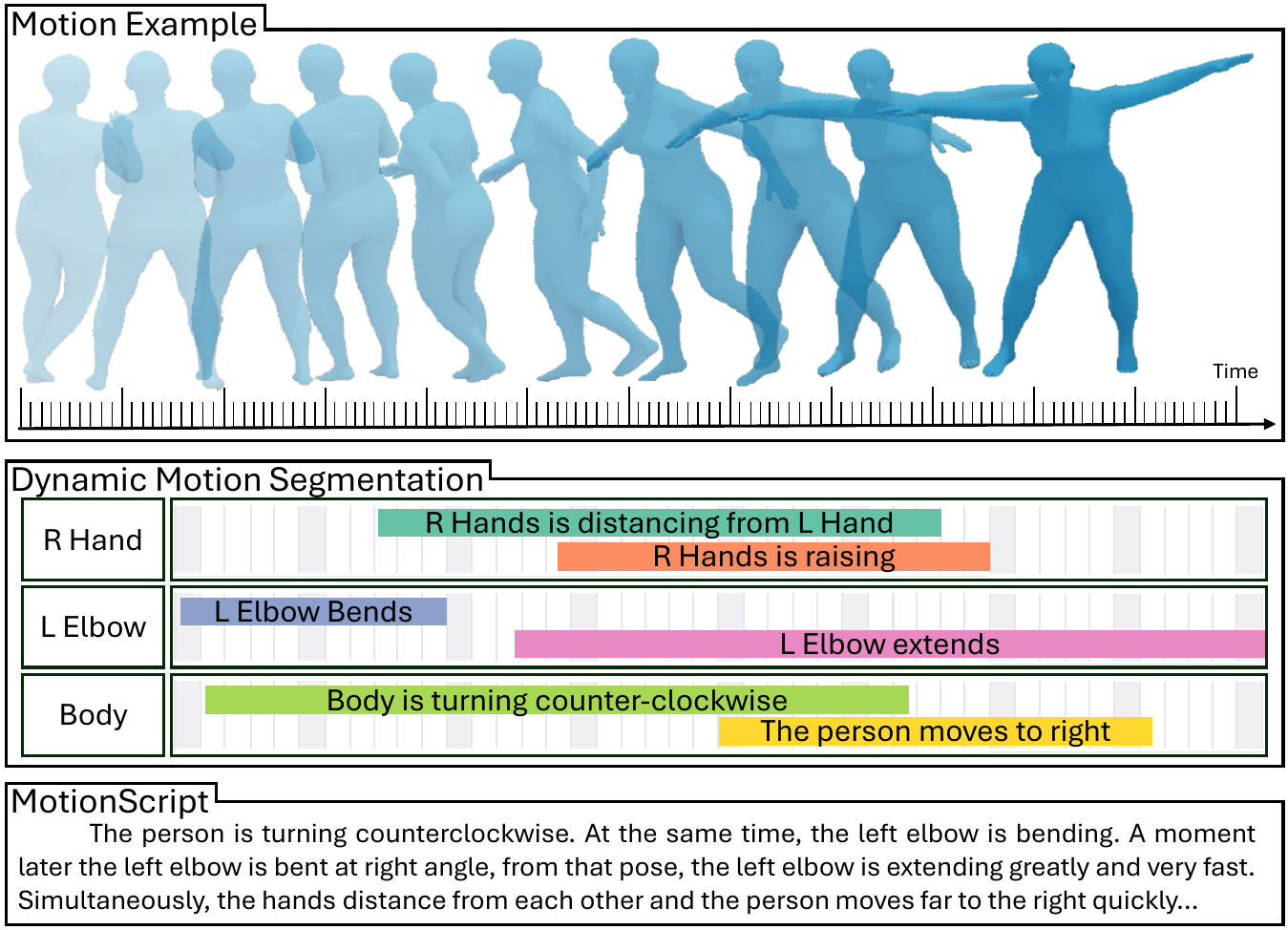} 
    \caption{Example motion sequence, dynamic motion segmentation with detected MotionCodes (Algorithm 1), and the resulting MotionScript, a structured motion description. 
    }
    \label{fig:caption-example}
\end{figure}
Furthermore, we define five types of elementary \textit{motioncodes} - angular, proximity, spatial relation, displacement, and rotation - based on relevant \textit{posecodes}.

    \noindent\textbf{Angular Motioncodes} describe joint movements, such as those of the elbows and knees, using angle posecodes.
    These motioncodes classify movement as `bending' or `extending', depending on the ${\mathbf{sgn}}({\mathbf{M}_{S}})$, and the intensity $\left| {\mathbf{M}_{S}} \right|$ into \{`significant', `moderate', `slight', `stationary'\}.


    \noindent\textbf{Proximity Motioncodes} track changes in the distance between two joints, e.g., hands. The direction of change, indicated by ${\mathbf{sgn}}({\mathbf{M}_{S}})$, is `spreading' for positive and `closing' for negative values. The intensity of the change is classed as \{`significant', moderate', slight', stationary'\} based on $\left| {\mathbf{M}_{S}} \right|$.

    \noindent\textbf{Spatial Relation Motioncodes} assess the relative movement between two joints, e.g., the movement of the \textit{``left hand from behind to in front of the head''}. 
    
    This motioncode categorizes movements into two directions regarding ${\mathbf{sgn}}({\mathbf{M}_{S}})$, with categories {`right-to-left', `left-to-right'}, {`behind-to-front', `front-to-behind'}, and {`above-to-below', `below-to-above'} corresponding to the X, Y, and Z axes, respectively. Since $\left| {\mathbf{M}_{S}} \right|$ only takes values of $-1$ or $+1$, no predefined intensity classes are assigned.


    \noindent\textbf{Rotation Motioncodes} determine the body's root rotational from the changes within orientation \textit{posecode} classes. Hence, ${{\mathbf{sgn}}(\mathbf{M}_{S})}$ is categorized into \{`leaning backward', `leaning forward'\} for the X-axis, \{`leaning right', `leaning left'\} for the Y-axis , and \{`turning clockwise, `turning counter-clockwise'\} for the Z-axis. The intensity $\left| {\mathbf{M}_{S}} \right|$ is classified into categories ranging from `significant' to `stationary'.

    \noindent\textbf{Displacement Motioncodes} extend position posecodes to examine the trajectory of the body's root and joints within 3D space. The ${\mathbf{sgn}}({\mathbf{M}_{S}})$, categorizes the \textit{motioncode} to \{`leftward', `rightward'\} along the X-axis, \{`upward', `downward'\} along the Y-axis, and \{`backward', `forward'\} along the Z-axis. This motioncode intensity provides insight on how the root body as well as joints traverse space. Section \ref{Subject_Selection} details the use of displacement motioncodes to identify the subject joint in symmetrical motions e.g., proximity \textit{motioncodes}.

\textbf{Motioncode Extraction Methodology:}
In order to detect dynamic segments within a posecode category over time that are non-stationary and exhibit a minimum motion, we propose the Dynamic Motion Segmentation algorithm, detailed in Algorithm \ref{algorithm_motion_detection}. This algorithm is robust against minor motions by treating transitions between adjacent categories as negligible movements if they fall below a predefined numbers of transitions, ensuring that the \textit{motioncodes} accurately capture the details of the underlying motions.
\begin{algorithm}[t]
\footnotesize
\caption{Dynamic Motion Segmentation }
\label{algorithm_motion_detection}
\begin{algorithmic}[1]
\State \textbf{Input}:
    \Statex - Posecode Sequence: Numerical values representing pose categories.
    \Statex - Maximum Range: Maximum segment length
\vspace{+1mm}

\Procedure{DetectMotions}{Posecode Sequence, Maximum Range}
    \For{each pose in Posecode Sequence}
        \State Identify changes in \textbf{posecode} categories (positive or negative direction).
        \State Merge adjacent same-direction poses into one motion segment.
        \State Calculate motion attributes i.e. spatial and temproal for each segment.
        \State Store detected motions with their parameters.
    \EndFor
    \State \textbf{return} Detected Motions
\EndProcedure
\end{algorithmic}
\end{algorithm}


\subsection{Motioncode Selection}
In this step, we aim to identify the most informative subset of the extracted \textit{motioncodes}. 
To compress the motion description, inspired by \cite{posescript}, we refine motion descriptions by eliminating non-discriminative or redundant spatial and temporal attributes based on statistical analysis.
For example, ``The left elbow slightly draws toward the right elbow'' is not discriminative enough due to the low intensity of $\left| {\mathbf{M}_{S}} \right|$. We also mark some of predefined motioncode attributes such as `significant bending' or `very fast' as rare to prioritize their inclusion. 


\subsection{Motioncode Aggregation}

At this step, we merge motioncodes together, if possible, to reduce the number of motioncodes and the overall output description length. We introduce a binning strategy based on the temporal attributes $\mathbf{M}_T$ such that each motioncode is assigned into one bin of fixed time interval of length $T_w$. A motioncode $\mathbf{M}$ is placed in the $n^{th}$ bin if $nT_w \leq \mathbf{M}{T_s} < (n+1)T_w$, treating all motioncodes within the same bin as concurrent events. This approach minimizes redundancy and shortens the overall output description.
Thus, we merge simultaneous \textit{motioncodes} using the following rules, which are applied randomly, enhancing diversity and coverage of all possible scenarios:

    \noindent\textbf{Entity-Based Aggregation} merges motioncodes with the same spatial attribute ${\mathbf{M}_{S}}$ but involving different joints of a larger entity. For instance, ``The left elbow gets close to the right foot'' and ``The left hand gets closer to the right foot'' can be combined as ``The left arm gets closer to the right foot.'' This allows joints that are closely related to be described as the motion of a larger entity.

    \noindent\textbf{Symmetry-Based Aggregation} combines motioncodes with identical ${\mathbf{M}_{S}}$ for symmetric joints on opposite sides of the body. For instance, ``the left elbow  bends" and ``the right elbow bends" aggregate to ``the elbows bend.''


Next, we apply keypoint-based and interpretation-based aggregation through time bins as follows.

    \noindent\textbf{Keypoint-Based Aggregation} merges motioncodes that share the same joint set but perform distinct actions. This process factors common key points as the subject and aggregates motioncodes both inside a bin and across adjacent bins within a specific range. The subject may be referred as ``it" or ``they" in the caption while explaining several aggregated motioncodes on that joint set.  
    For example, ``the right elbow bends significantly" and ``spreads out from the left elbow" are in the same bin, while ``extends slightly" is in the subsequent bin. These are aggregated as the ``right elbow bends and spreads out from the other elbow. Afterward, it extends slightly" with ``afterward" indicating the chronological order of motions discussed in Sec. \ref{Section_timecode}.

    \noindent\textbf{Interpretation-based Aggregation} fuses the motioncodes with the same spatial attribute ${\mathbf{M}_{S}}$ but operating on different joints. For instance, ``the left elbow bends slightly" and ``the right knee bends slightly" combine to ``the left elbow and right knee bend slightly". We apply keypoint and interpretation-based aggregation within a range spanning $T_{range}$ bins before and after, including simultaneous motions within the same bin. The time relation between aggregated motions is maintained to preserve the chronological sequence of actions.

\noindent\textbf{Timecode-Based Aggregation}
\label{Section_timecode} supports the temporal relationship between \textit{motioncodes} or their chronological order.
For instance, consider two \textit{motioncodes} ``the left elbow bends slightly" followed by ``the right knee spreads out from the other one". Therefore, merging them while considering their temporal relation would be ``the left elbow bends and immediately after, the right knee spreads out from the other one". \\
Since each aggregated \textit{motioncode} spans multiple time bins, it's crucial to preserve their chronological order. For instance, consider ``the right elbow bends" and ``the right elbow extends" are in the $n$ and $n+5$ bins respectively, and ``the right knee bends" is in the $n+1$ bin. The aggregated description for the right elbow might be ``the right elbow bends and a few seconds later, it extends'' which exceeds the bin for the right knee \textit{motioncodes}. Therefore, we use ``A moment before, the right knee bends" to reflect the chronological order of \textit{motioncodes} in the description.

\subsection{Motioncode Conversion}
As the last step, we plug the \textit{motioncodes} elements into a randomly selected sentence template. These templates represent the dynamic motions, integrating \textit{motioncodes} attributes and their chronological order. Template components, i.e. verbs and spatial/temporal adjectives, come from a broad dictionary for each category. We also implement a strategy for subject detection and pose injection to improve caption accuracy, as described in the following sections.


\subsubsection{Subject Selection} 

\label{Subject_Selection}
{
For symmetrical \textit{motioncodes}, such as proximity, we identify the most active joint as the subject when contributions are uneven based on a predefined threshold.
To analyze displacement \textit{motioncodes}, we calculate the Euclidean distance for each joint \( J \) as:
\vspace{-1.5mm}
\[
d = \sqrt{(x_{T_e} - x_{T_s})^2 + (y_{T_e} - y_{T_s})^2 + (z_{T_e} - z_{T_s})^2}
\]
where \((x_{T_s}, y_{T_s}, z_{T_s})\) and \((x_{T_e}, y_{T_e}, z_{T_e})\) represent the 3D coordinates of joint \( J \) at the start time \( T_s \) and end time \( T_e \), respectively.
If a specific joint, such as the left hand, contributes more than a certain threshold (e.g., 60\%) to the motion, it is identified as the subject (e.g., \emph{``the left hand moves away from the right hand.''}). Otherwise, we use phrases such as \emph{``each other''} or \emph{``one another''} for both joints (e.g., \emph{``the left hand and right knee move away quickly from each other.''}).}

\subsubsection{Pose Injection} 
\label{pose_injection}
We inject the initial pose state and/or final state of \textit{posecodes} that are relevant to the joints and the type of the \textit{motioncodes}. For instance, if the \textit{motioncode} is \textit{``the left elbow bends slightly''}, we also add the angle \textit{posecodes} description \textit{``the left elbow extends completely''}. However, relevant \textit{posecodes} selection depends on their eligibility within the posescript process. There are instances where a \textit{posecode} may not be eligible, and at times, may require a larger set to cover all necessary \textit{posecodes} due to aggregation. 
For instance, the \textit{posecode} for the left elbow might aggregate to \textit{``both elbows bend completely''}, which extends beyond the targeted joint.
We then use a modified weighted set cover algorithm \cite{chvatal1979greedy} to find the maximum covering set of associated \textit{posecodes} with the minimum number of irrelevant \textit{posecodes}. 
Finally, we inject the pose description into the motion description by blending it with transitions selected randomly from the Pose-to-Motion or Motion-to-Pose templates.  e.g., \textit{\{comma, period, `and', `from this position', `leads to', ...\}}.

\begin{figure}[t]
    \centering
    \includegraphics[width=0.49\textwidth]{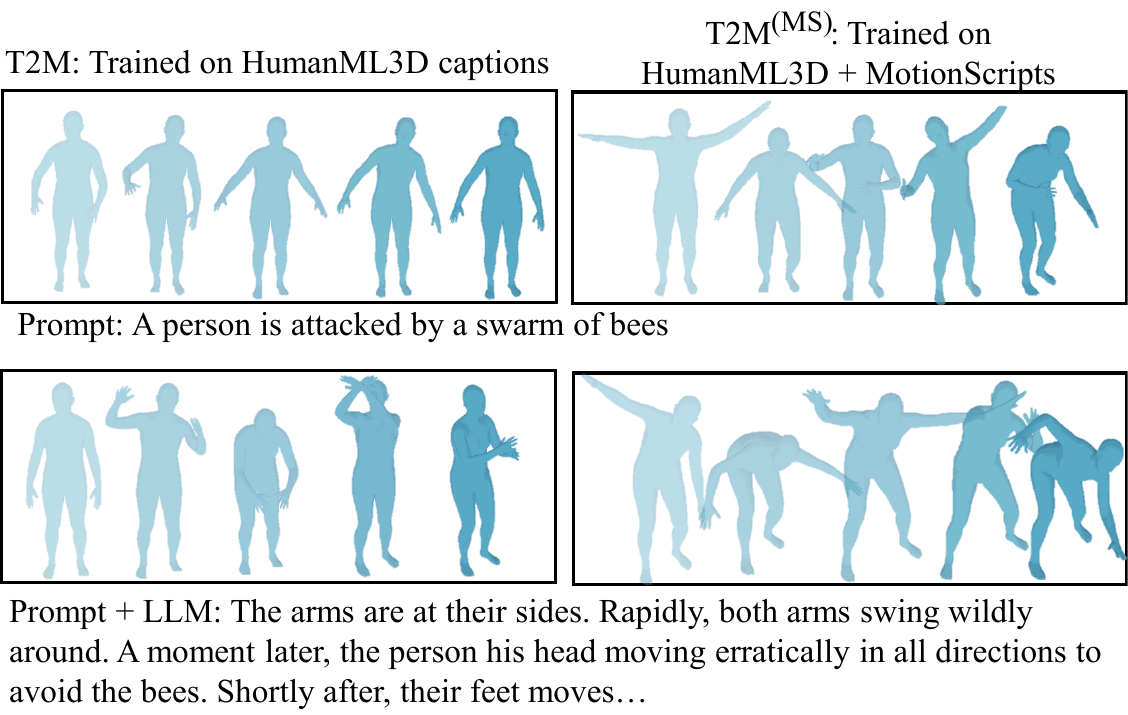} 
    \label{fig:Rachel_3}
    \caption{Comparison of motion generation from T2M trained on human annotations (left) and T2M\textsuperscript{\textbf{(MS)}} trained on MotionScript-augmented data (right), using plain text (top) and LLM-enhanced prompts (bottom).}
    \vspace{-6mm}
\end{figure}

\section{Application: Bridging the gap between LLMs and Motions}
\label{sec:experiment}
An exciting application of MotionScript is its ability to interface with LLMs as a translation layer between high-level, out-of-distribution human motion descriptions and our structured MotionScript format. By converting arbitrary motion descriptions—such as \textit{“while following a treasure map, you suddenly find the treasure”} or \textit{“a person pretending to be an eagle”}—into detailed MotionScript captions, LLMs enable a powerful system for generating precise 3D motions from diverse inputs. We prompt an LLM with examples of MotionScript captions to produce MotionScript-like descriptions of the intended motion, as shown in Fig. \ref{fig:experiment_app_llm}, with additional examples in the demo video and \href{https://pjyazdian.github.io/MotionScript}{project webpage}.

\subsection{Out of Distribution Text-to-Motion (T2M) Generation}
    In this section, we study whether motion generation from unseen text prompts can be improved by augmenting T2M training data with MotionScript's fine-grained captions (Fig.~\ref{fig:caption-example}). We build upon previous work such as T2M-GPT \cite{T2MGPT}, which generates motions from text descriptions by training on the {HumanML3D} dataset. HumanML3D is a widely used dataset that merges the AMASS \cite{AMASS} and HumanAct12 \cite{Action2motion} datasets and guarantees that each motion sample is paired with at least three captions. It provides 28.59 hours of motion data across 1,461 motion sequences, each with a maximum duration of 10 seconds, and 44,970 textual descriptions, 12 words on average. 

One challenge in evaluating out-of-distribution data is that groundtruth data may not exist. Typically, groundtruth text + motion pairs are used to calculate quantitative metrics such as R-Precision, FID, Diversity, and Multimodality, but we aim to evaluate on text that does not have an existing motion pair. These metrics are also limited in that they may not fully reflect aspects of semantic appropriateness \cite{GENEA2023, Make-An-Animation, Como}. For instance, MAA and CoMo received more favorable user reviews despite lower R-Precision and FID, and some methods even surpass ground truth in R-P@1. To address this, we conduct human perception studies as the gold standard for evaluating the quality of text-to-motion generation.

\subsection{Experimental Setup}

We conducted experiments using combinations of OOD captions and models trained with augmented data, as follows.

\textbf{Caption Test Data.} As testing data, we curated a set of 44 out-of-distribution captions, which we call \emph{plain} captions. These captions were derived from miming and improv acting exercises\footnote{https://www.forteachersforstudents.com.au/site/wp-content/uploads/pdfs/tmpl-mime-activity-ideas.pdf}. Example captions are shown in Table \ref{tab:best_captions} and the complete set can be found on the \href{https://pjyazdian.github.io/MotionScript}{project webpage}. In addition, we created \emph{detailed} captions, which are \emph{plain} captions converted to a more detailed version of the same text, using ChatGPT. We prompted  ChatGPT-3.5-turbo \cite{openai2023chatgpt} to either produce detailed captions by prompting it with \emph{``Describe a person's body movements while performing the action \{X\} in detail. Please respond 2-4 sentences."} \textbf{(Detailed-LLM)}, or with examples of MotionScript such as \emph{``Describe a person's body movements while performing the action \{X\} in detail. Please respond 2-4 sentences in the style of following examples: \{motionscript-examples\}"} (see exact prompt on project webpage) \textbf{(Detailed-MS)}.

\textbf{Trained models.} We trained T2M-GPT\cite{t2m-gpt} with 3 different datasets, resulting in models as follows:
 
 \begin{enumerate}
\item{\textbf{T2M (Baseline)}. T2M-GPT \cite{t2m-gpt} architecture trained on the original HumanML3D dataset}
 \item{\textbf{T2M\textsuperscript{\textbf{(LLM)}}(Baseline)}. T2M-GPT trained on HumanML3D plus ChatGPT-augmented training data Detailed-LLM, described above.} 
\item{\textbf{T2M\textsuperscript{\textbf{(MS)}} (Proposed)}. T2M-GPT trained on HumanML3D plus MotionScript-augmented training data Detailed-MS, described above.}
 \end{enumerate}

    \begin{table}[t]
    \centering
   
    \begin{tabular}{|l|c|c|c|}
    \hline
    \textbf{Model} & \textbf{Training Data} & \textbf{Exp. 1} & \textbf{Exp. 2} \\
    \hline
    T2M & HumanML3D & \checkmark &\\
        T2M\textsuperscript{(LLM)} & HumanML3D + LLM Aug. &  & \checkmark \\
    T2M\textsuperscript{(MS)} & HumanML3D + MotionScript Aug.& \checkmark & \checkmark\\

    \hline
    \end{tabular}
     \caption{Training datasets used in Exp. 1 and Exp. 2. Exp. 1 compares the baseline model T2M with MotionScript-augmented training T2M\textsuperscript{(MS)}. Exp. 2 compares T2M\textsuperscript{(MS)} with LLM-augmented training (T2M\textsuperscript{(LLM)}).}
    \label{tab:training_comparison}
    
    \end{table}
    
We conducted two experiments to validate our approach for text-aligned human motion generation, focusing on evaluating the effectiveness of different training strategies in open-vocabulary, out-of-distribution scenarios. Table~\ref{tab:training_comparison} summarizes the model training data for each experiment.

\subsection{Experiment 1: MotionScript Augmentation}

The purpose of this experiment was to understand the effect of training T2M-GPT on the dataset with and without MotionScript augmentation, as a first step to evaluate the potential of MotionsScript to improve motion quality and generalization. In this initial study, we randomly selected 20 prompts from our caption test dataset.
      
We compared motions generated from plain or detailed captions using T2M and T2M\textsuperscript{\textbf{(MS)}} as shown in Table II. For each caption, participants were asked, ``Rank the four videos below based on how well it fits the caption." and shown four motion clips (a), (b), (c), and (d). Participants could adjust their rankings as needed, and the order of clips was shuffled. Twenty-three participants completed the study and the experiment lasted approximately 20 minutes on average. 

We completed Chi Squared Goodness of Fit tests, followed by confidence intervals (CIs) to assess preference for specific motion generation methods. We found a significant preference at $\alpha = 0.01$ for our proposed method Detailed-MS + T2M\textsuperscript{\textbf{(MS)}}, most often selected as the top preferred motion (count=149), $\chi^{2}(3, N=460)=14.47, p=.002, CI=25-36\%$, followed by Detailed-MS + T2M (count=110), Plain + T2M\textsuperscript{\textbf{(MS)}} (count=110) and Plain + T2M (count=82). This confirms that our method provides better alignment than baseline models, highlighting how MotionScript enhances existing methods and bridges the gap between LLMs and motion representation. Table~\ref{table:experiment_humanstudy_first} presents the Chi-Squared statistics, showing a significant preference for MotionScript as first choice.

\begin{table}[t]
\centering
\begin{tabular}{|l|l|c|c|c|}
\hline
\textbf{Caption Type} & \textbf{Model} & \textbf{1st Choice Ct.} & $\chi^2$ & p-value \\
\hline
(a) Plain & T2M & 82 & 25.22 & $< 0.0001$* \\
(b) Plain & T2M\textsuperscript{\textbf{(LLM)}} & 110 & 20.23 & $< 0.0001$* \\
(c) Detailed-MS & T2M & 116 & 7.84 & 0.049 \\
(d) Detailed-MS & T2M\textsuperscript{\textbf{(MS)}} & \textbf{149} & 40.83 & $< 0.0001$* \\
\hline
\end{tabular}
\caption{Experiment 1. Chi-Squared preference result with first choice counts. We observe a significant preference of T2M\textsuperscript{\textbf{(MS)}} over the baselines.}
\label{table:experiment_humanstudy_first}
\vspace{-4mm}
\end{table}

\subsection{Experiment 2: LLM Text Augmentation}
  
The promising results of Experiment 1 led us to explore a more difficult baseline. In this second experiment, we compared our best result from Exp. 1, Detailed-MS captions input into T2M-GPT trained with MotionScript-augmented data (T2M\textsuperscript{\textbf{(MS)}}), versus Detailed-LLM captions as input~\cite{Action-gpt} into T2M-GPT trained with LLM-augmented data (T2M\textsuperscript{\textbf{(LLM)}}).




    \label{sec_User_study}

We randomly selected 34 plain prompts from our set of 44 captions along with their detailed captions, allowing us to compare Detailed-MS + T2M\textsuperscript{\textbf{(MS)}} with Detailed-LLM +  T2M\textsuperscript{\textbf{(LLM)}} while maintaining consistency with the trained data for each model. For each caption, participants were asked, ``How well does Video X match the prompt?" and provided a rating on a Likert scale from 1 (doesn't match at all) to 7 (matches extremely well). In addition, they selected their preferred motion for the prompt from the two options. The order of clips was shuffled, and participants could adjust their ratings as needed. Thirty participants recruited via Prolific.com (inclusion criteria: 18+ years old, can read and comprehend English) completed the survey, and the average duration of this experiment was approximately 25 minutes. 

The results showed a strong preference for Detailed-MS + T2M\textsuperscript{\textbf{(MS)}} over Detailed-LLM + T2M\textsuperscript{\textbf{(LLM)}}. Detailed-MS + T2M\textsuperscript{\textbf{(MS)}} was preferred in 82\% of all n=1817 responses. Likert ratings results from this study are shown in Fig. \ref{fig:experiment_humanstudy_Qs}. We can observe a general trend with our proposed method achieving a higher Likert score, with a significantly higher score for 7 captions in particular, which we report in Table \ref{tab:best_captions}. 
One potential cause of failure cases may be the insertion of parasitic motions in T2M\textsuperscript{\textbf{(MS)}}, possibly due to hallucinated MotionScript generated by the LLM.
Overall, through human studies, our experiments confirmed that MotionScript improves text-to-motion generation using our proposed setup.

\begin{table}[]
    \centering
    \begin{tabularx}{\linewidth}{X}
    \hline
    Q7. You are an eagle flying through the air. \\
    Q9. You are walking along but your dog keeps grabbing hold of your slipper. \\
    Q15. You meet your child at the airport after a long separation and give them a big hug. \\
    Q17. While following a treasure map, you suddenly find the treasure. \\
    Q19. You are raking up leaves but they are falling off the tree faster than you can gather them. \\
    Q21. You are taking a shower, when suddenly the water goes cold. \\
    Q25. You are sitting in a lecture and fighting off sleep. \\
    \hline
    \end{tabularx}
    \caption{Experiment 2. Examples of OOD plain captions.}
    \label{tab:best_captions}
    \vspace{-6mm}
\end{table}

        
\begin{figure}[t]
    \centering
    \includegraphics[width=0.9\columnwidth]{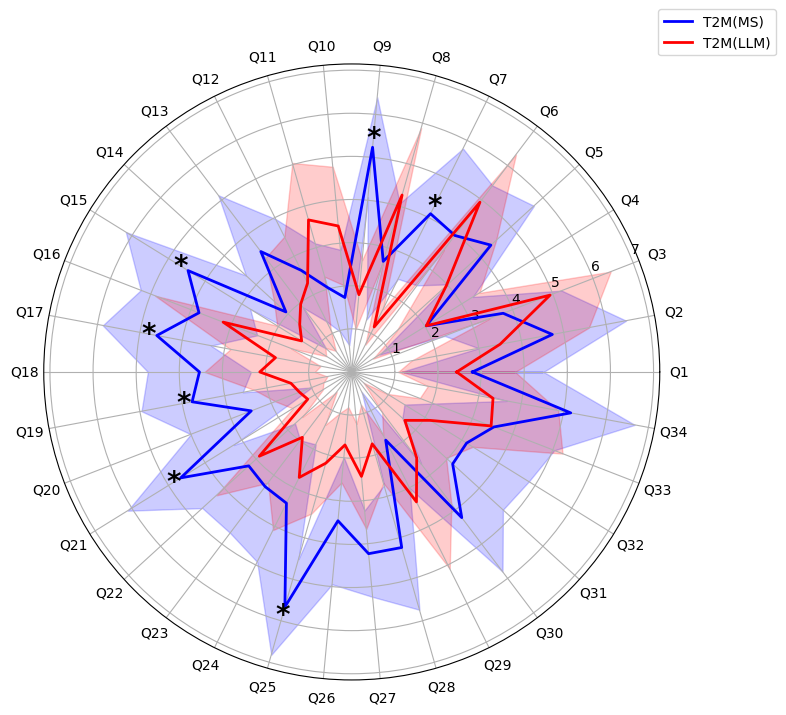}
    \caption{
        Experiment 2 results comparing T2M\textsuperscript{\textbf{(MS)}} (blue) and T2M\textsuperscript{\textbf{(LLM)}} (red). We show mean ratings with standard deviation across 34 evaluation questions from 1 (worst) to 7 (best). MotionScript (blue) shows a trend in outperforming the baseline, with significant differences marked with a star (*).
    }
    \vspace{-5mm}
    \label{fig:experiment_humanstudy_Qs}
\end{figure}

\section{Conclusion and Future work}
\label{sec:conclusion}

In this work, we introduced MotionScript, a novel semantic representation for 3D human motion that links fine-grained spatial and temporal aspects to natural language descriptions. Unlike traditional captioning methods, MotionScript offers an interpretable mapping between 3D motion and text, enabling large language models to generate 3D motions from high-level descriptions, even for out-of-distribution samples. A human study confirms the effectiveness of our approach.
Future work will expand MotionScript to include additional motion features, such as fine motor gestures, facial expressions, and gaze shifts, and compare it to other motion augmentation methods. To further improve the readability of the generated descriptions, which are structured but sometimes complex, LLM-based revision could be explored to produce more concise and human-like captions. We also aim to explore its potential in enhancing multimodal AI systems that integrate language, vision, and motion.







\bibliographystyle{IEEEtran}
\bibliography{IEEEabrv}

\end{document}